\documentclass[review]{elsarticle}

\usepackage{lineno,hyperref}
\hypersetup{pdfauthor=author}

\modulolinenumbers[5]

\journal{Expert Systems With Applications}
\usepackage[T1]{fontenc}

\usepackage{times}
\usepackage{latexsym}
\usepackage{microtype}
\usepackage{graphicx}
\usepackage{subfigure}
\usepackage{amsmath}
\usepackage{amssymb}
\usepackage{threeparttable}
\usepackage{booktabs}
\usepackage{multirow}
\usepackage{pifont}
\usepackage{bm}
\usepackage{tabularx}

\usepackage{color}

\let\oldequation\equation
\let\oldendequation\endequation

\renewenvironment{equation}
  {\linenomathNonumbers\oldequation}
  {\oldendequation\endlinenomath}
  
\usepackage{soul}
\usepackage{color,xcolor}
\definecolor{blue}{RGB}{81, 193, 200}
\definecolor{pink}{RGB}{233, 98, 121}

\biboptions{authoryear}

\begin{document}

\begin{frontmatter}

\title{Unlock the Potential of Counterfactually-Augmented Data in Out-Of-Distribution Generalization}

%% Group authors per affiliation:
\author[address1]{Caoyun Fan} \ead{fcy3649@sjtu.edu.cn}
\author[address2]{Wenqing Chen} \ead{chenwq95@mail.sysu.edu.cn}
\author[address1]{Jidong Tian} \ead{frank92@sjtu.edu.cn}
\author[address1]{Yitian Li} \ead{yitian_li@sjtu.edu.cn}
\author[address1]{Hao He\corref{correspondingauthor}} \ead{hehao@sjtu.edu.cn}
\author[address1]{Yaohui Jin} \ead{jinyh@sjtu.edu.cn}

\renewcommand{\thefootnote}{\fnsymbol{footnote}}

\cortext[correspondingauthor]{Corresponding author. }

%% or include affiliations in footnotes:
\address[address1]{MoE Key Lab of Artificial Intelligence, AI Institute, Shanghai Jiao Tong University, Shanghai, China}
\address[address2]{School of Software Engineering, Sun Yat-sen University, Guangzhou, China}

\begin{abstract}

Counterfactually-Augmented Data (CAD) -- minimal editing of sentences to flip the corresponding labels -- has the potential to improve the Out-Of-Distribution (OOD) generalization capability of language models, as CAD induces language models to exploit domain-independent causal features and exclude spurious correlations. However, the empirical results of CAD's OOD generalization are not as efficient as anticipated. In this study, we attribute the inefficiency to the myopia phenomenon caused by CAD: language models only focus on causal features that are edited in the augmentation operation and exclude other non-edited causal features. Therefore, the potential of CAD is not fully exploited. To address this issue, we analyze the myopia phenomenon in feature space from the perspective of Fisher's Linear Discriminant, then we introduce two additional constraints based on CAD's structural properties (dataset-level and sentence-level) to help language models extract more complete causal features in CAD, thereby mitigating the myopia phenomenon and improving OOD generalization capability. We evaluate our method on two tasks: Sentiment Analysis and Natural Language Inference, and the experimental results demonstrate that our method could unlock the potential of CAD and improve the OOD generalization performance of language models by 1.0\% to 5.9\%. % \uncertain{ok}

\end{abstract}

\begin{keyword}
Counterfactually-Augmented Data; Out-Of-Distribution Generalization; Feature Extraction; Language Models % % \uncertain{ok}
\end{keyword}

\end{frontmatter}

% \linenumbers

\section{Introduction}
\label{s1}

Out-Of-Distribution (OOD) generalization \citep{Shen2021TowardsOG,Wang2021GeneralizingTU} refers to the capability of machine learning models to make accurate predictions on data that is outside of its training data distribution. OOD generalization is crucial because it ensures that the machine learning models perform reliably in complex real-world scenarios \citep{DBLP:conf/iclr/NagarajanAN21,DBLP:conf/iclr/WilesGSRKDC22}. However, despite the remarkable performance of language models in Natural Language Processing (NLP) \citep{Devlin2019BERTPO,Liu2019RoBERTaAR,DBLP:conf/nips/BrownMRSKDNSSAA20}, the OOD generalization capability of language models is always less than ideal \citep{DBLP:conf/naacl/SenS0A22,QuioneroCandela2009DatasetSI}. Many studies \citep{Teney2020LearningWM,Joshi2021AnIO} pointed out that such limited generalization capability partly arises from the language models' exploitation of spurious correlations \citep{Mitchell2007TheNF,Torralba2011UnbiasedLA,McCoy2019RightFT,Wang2020IdentifyingSC} in the dataset \citep{DBLP:journals/widm/NtoutsiFGINVRTP20}. Specifically, the language models tend to exploit dataset-specific correlation bias \citep{Teney2020LearningWM,Spears2004EvaluationAS} rather than the intrinsic properties of tasks \citep{DBLP:journals/coling/FederOSR21,DBLP:journals/eswa/BueffCCJRMB22} to make predictions. However, the dataset-specific correlations are unreliable and could not be generalized well to OOD scenarios. 
% For example, in the NLI task, language models highly rely on unintended lexical features \citep{DBLP:conf/acl/NivenK19,DBLP:conf/naacl/DuMJDDGSH21} (e.g. negation words) to determine relationships between sentences. 

\begin{figure}[ht]
    \centering
    \includegraphics[width=0.7\columnwidth]{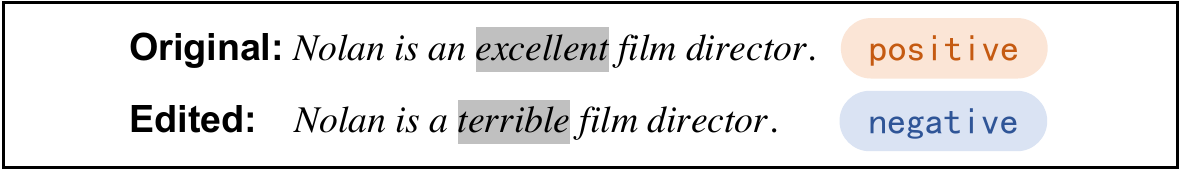}
    % \vspace{-0.6cm}
    \caption{An example of constructing CAD in the Sentiment Analysis task. We expect CAD to induce language models to exploit the causal features (in grey) and exclude the possible spurious correlations (e.g., \emph{Nolan} in the sentence). }
    \label{f1-1}
    % \vspace{-0.3cm}
\end{figure}

To solve the problem of spurious correlations in OOD generalization, a recent promising direction is Counterfactually-Augmented Data (CAD) \citep{Kaushik2020LearningTD,Wang2021RobustnessTS,Yang2021ExploringTE}: minimal editing of the sentence to flip the corresponding label $Y$, where the edited part is considered to be the task's intrinsic properties and have a causal effect on the label (as shown in Fig. \ref{f1-1}). In this paper, we refer to the pairs consisting of the original sentence and the corresponding edited sentence as counterfactual sentence pairs. Unlike the Independent Identically Distribution (IID) principle of most data augmentation methods \citep{DBLP:journals/jbd/ShortenKF21a}, CAD aims to deliberately change the data distribution of the dataset so that the language models could spontaneously distinguish between intrinsic properties of tasks and dataset-specific bias, and then exclude spurious correlations. Specifically, in a counterfactual sentence pair, CAD attempts to keep the correlated features $h_r$ constant while the causal features $h_c$ change through the minimal editing\footnote{Under the ideal conditions \citep{Joshi2021AnIO}, each sentence consists of the causal features $h_c$ whose joint distribution with labels is invariant, and the correlated features $h_r$ whose joint distribution would vary. }. Then, the classifier $\Phi$ could make predictions spontaneously based on the causal features and then exclude the interference of the correlated features as: 
\begin{equation}
    \begin{split}
        \Phi(h_c, h_r) &= Y, \\
        \Phi(h_c^*, h_r) &= Y^*, 
    \end{split}
    \label{e1-1}
\end{equation}

\noindent where $h_c^*$ and $Y^*$ denote the causal features and the label of the edited sentence in the counterfactual sentence pair, respectively. Intuitively, in Eq. \ref{e1-1}, the classifier $\Phi$ no longer focuses on $h_r$ because different labels correspond to the same $h_r$. Based on this assumption, we analyze the effectiveness of CAD from the perspective of Fisher's Linear Discriminant \citep{Fisher1936THEUO}, which becomes the basis for our understanding of the OOD generalization potential of CAD. However, extensive experiments \citep{Khashabi2020MoreBF,DBLP:conf/naacl/SenS0A22} demonstrated that CAD is not efficient in improving language models' OOD generalization capability, especially in more complex tasks \citep{Huang2020CounterfactuallyAugmentedST}. This is not in line with our expectations for CAD. 

In this work, we attribute the inefficiency of CAD in OOD generalization to the CAD-imposed myopia phenomenon: language models focus only on the causal features edited by counterfactual augmentation operation, which means the correlated features along with other non-edited causal features are excluded. Further, the myopia phenomenon is also clearly presented in the setting of Fisher's Linear Discriminant. However, excluding only some causal features is not what we want, as \cite{Balashankar2021CanWI} indicated that all causal features are beneficial for OOD generalization. To \textbf{E}xtract more complete \textbf{C}ausal \textbf{F}eatures and unlock the potential of CAD for OOD generalization, we design the ECF algorithm: introducing additional constraints in the training process based on the structural properties of CAD. Specifically, guided by Fisher's Linear Discriminant, we extract the invariant causal features over both distributions of CAD and the original data by the Invariant Risk Minimization \citep{DBLP:journals/corr/abs-1907-02893} method (dataset level) and constrain the correlated feature similarity of counterfactual sentence pairs (sentence level). Through extensive experiments across multiple language models and NLP tasks, we conclude that the ECF algorithm could help language models to extract more complete causal features, and then improve the OOD generalization capability. 

This study is organized in the following structure: in Section \ref{s3}, we analyze the effectiveness of CAD from the perspective of Fisher's Linear Discriminant; in Section \ref{s4}, we theoretically explain the shortcomings of CAD in OOD generalization, which is called myopia phenomenon; in Section \ref{s5}, we introduce two additional constraints based on CAD’s structural properties to help language models extract more complete causal features; in Section \ref{s6} \& \ref{s7}, we confirm that our method can unlock the potential of CAD in OOD generalization through extensive experiments. We summarize our contributions as follows: 
\begin{itemize}
    \item We attribute the inefficiency of CAD in OOD generalization to the myopia phenomenon and 
    explicitly demonstrate this from the perspective of Fisher's Linear Discriminant. 
    \item To mitigate the myopia phenomenon and unlock the potential of CAD in OOD generalization, we design two additional constraints based on CAD's structural properties to help language models extract more complete causal features. 
    \item We evaluate our method on two NLP tasks, and the experimental results demonstrate that our method could successfully unlock the potential of CAD and improve the OOD generalization performance of language models by 1.0\% to 5.9\%. % \uncertain{ok}
\end{itemize}

\section{Related Work}
\label{s2}

Our study aims to improve the OOD generalization of language models using augmented text dataset (CAD). Therefore, we introduce the related work from two perspectives: text data augmentation and OOD generalization. 

\subsection{Text Data Augmentation}
\label{s2-1}

Text data augmentation \citep{DBLP:journals/csur/BayerKR23} is a highly important research area since the NLP community has a great need for quantity and quality of data \citep{DBLP:conf/emnlp/KarimiR021}. \cite{DBLP:journals/jbd/ShortenKF21a} categorized text augmentation methods as symbolic augmentation and neural augmentation. Symbolic augmentation does not involve neural networks but is based on several symbolic rules \citep{DBLP:journals/eswa/HaralabopoulosT21} (e.g., random swapping, random deletion, random insertion and random synonym substitution) to increase the diversity of texts. Representative studies include EDA \citep{DBLP:conf/emnlp/WeiZ19}, AEDA \citep{DBLP:conf/emnlp/KarimiR021}, etc. Neural augmentation relies on auxiliary neural networks to generate additional data. For example, \cite{DBLP:conf/ecir/Kohli21} employed a translation model to perform round-trip translations of texts to obtain paraphrased texts, \cite{DBLP:conf/emnlp/NgCG20,DBLP:journals/eswa/FengZZM22} utilized the language model to replace the masked words to produce diverse texts. However, Most of these augmentation methods cannot significantly change the data distribution. 

CAD \citep{Kaushik2020LearningTD} is an emerging technique in text data augmentation, which aims to help language models extract causal features by changing the data distribution. The construction of CAD is relatively challenging. Initially, \cite{Kaushik2020LearningTD,Khashabi2020MoreBF} used manual annotation to obtain high-quality CAD, although accompanied by high costs and inefficiencies. Subsequently, several automatic generation methods were proposed: \cite{Wang2021RobustnessTS} substituted causal words using a statistical matching approach; \cite{Yang2021ExploringTE} employed pre-trained models to control the minimal edits; \cite{Chen2021ReinforcedCD} discriminated the generated counterfactual sentence by reinforcement learning; \cite{DBLP:conf/acl/ChenGBS023} utilized large language models to generate CAD in a few-shot manner.

However, CAD did not perform efficiently in OOD generalization according to the empirical experimental results \citep{Huang2020CounterfactuallyAugmentedST,Khashabi2020MoreBF}, and \cite{Joshi2021AnIO} attributed this inefficiency to the lack of diversity in counterfactual editing and demonstrated their view in terms of theoretical explanations and experimental results. Therefore, it has become a research problem to improve the efficiency of CAD. The previous studies could be divided into two categories: improving the augmentation quality of counterfactual sentence pairs \citep{Wang2021RobustnessTS,Yang2021ExploringTE}; debiasing for specific bias \citep{Balashankar2021CanWI,DBLP:conf/birthday/LuMWAD20,DBLP:conf/acl/ZmigrodMWC19} (e.g., gender, race) in CAD. To the best of my knowledge, our study is the first attempt to improve the efficiency of CAD by designing additional constraints. Notably, our method does not require changing the dataset or obtaining further information about specific biases, which makes it applicable to a wider range of scenarios.  % \uncertain{ok}
% which neither change the dataset nor require additional information about specific biases, and is the most general application scenario. % \uncertain{ok}

\subsection{Out-Of-Distribution Generalization}
\label{s2-2}

Most existing machine learning models assumed that the training and test data would follow the Independent Identically Distribution (IID) principle \citep{DBLP:conf/nips/Vapnik91,DBLP:journals/tnn/Vapnik99}, yet many studies revealed the vulnerability of machine learning models when the data distributions changed \citep{QuioneroCandela2009DatasetSI}. Since the IID principle is difficult to be satisfied in real-world scenarios \citep{DBLP:conf/iclr/NagarajanAN21,DBLP:conf/iclr/WilesGSRKDC22}, the study of Out-Of-Distribution (OOD) generalization \citep{Wang2021GeneralizingTU} is of great urgency in the academic community as well as the industry community \citep{ASUTKAR2023119016}. 

Representation learning \citep{DBLP:journals/pami/BengioCV13,DBLP:journals/eswa/TongTCX21} is one of the main focuses of research in OOD generalization, and \cite{Shen2021TowardsOG} classified representation learning methods in OOD generalization into two main categories: domain-invariant representation learning and feature disentanglement. Domain-invariant representation learning aims to find features that remain invariant to different domains, and \cite{DBLP:conf/nips/Ben-DavidBCP06} has proven its effectiveness theoretically. Representative studies include Invariant Risk Minimization \citep{DBLP:journals/corr/abs-1907-02893}, Maximum Mean Discrepancy \citep{DBLP:journals/jmlr/GrettonBRSS12,DBLP:journals/tist/WangCFYHY20}, etc. The purpose of feature disentanglement \citep{DBLP:conf/icml/PiratlaNS20} is to decompose a feature representation into two understandable sub-features, with one feature being domain-invariant feature and the other being domain-specific. Representative studies include UndoBias \citep{DBLP:conf/eccv/KhoslaZMET12}, DIVA \citep{DBLP:conf/midl/IlseTLW20}, etc. CAD aims to help language models obtain generalizable and robust features, so we consider that CAD is also a type of representational learning. It is worth noting that unlike most previous representation learning methods, CAD \citep{Kaushik2020LearningTD,Wang2021RobustnessTS,Yang2021ExploringTE} attempts to obtain such robust features by changing the data distribution (rather than designing additional modules, adjusting  optimization objectives, etc.). 

Intuitively, CAD highlights the causal information in the dataset based on domain-specific knowledge \citep{Wang2021GeneralizingTU}, in order to help language models spontaneously extract robust features in the dataset. From the perspective of causal learning \citep{DBLP:journals/tkdd/YaoCLLGZ21}, causality is a more fine-grained description of dataset beyond statistics \citep{Pearl2009CausalII}, and machine learning models, on the other hand, are statistical models based on probability distributions \citep{Shah2020ThePO}, so they lack the capability to distinguish between correlation and causality in dataset \citep{DBLP:journals/coling/FederOSR21}. Therefore, CAD highlighting the causal information (and exclude correlation) can theoretically improve the OOD generalization of language models. However, our study points out that CAD is likely to fail to highlight all causal information, which is detrimental to OOD generalization, while our proposed method aims to help language models extract more complete causal information. 

% However, based on domain-specific knowledge \citep{Wang2021GeneralizingTU}, humans highlight the causal information in the dataset to construct CAD, so the data distribution of CAD enables language models to spontaneously utilize causality and exclude correlation to make predictions. % \uncertain{ok}
% Specifically, the researcher's motivation for proposing CAD is that humans could describe causal relationships in the dataset based on domain-specific knowledge \citep{Wang2021GeneralizingTU}, and highlight these causal relationships through data augmentation operations. 

\section{Effectiveness of CAD}
\label{s3}

In this section, we formalize the sentence features as a combination of Gaussian distributed features and explain the effectiveness of CAD from the perspective of Fisher's Linear Discriminant. % \uncertain{ok}

\subsection{Fisher's Linear Discriminant}
\label{s3-1}

% balanced
We assume that $\{X_i, Y_i\}_N$ is a class-balance binary classification dataset, where $N$ represents the number of samples. Without loss of generality, we assume that the mean value of samples is 0: 
\begin{equation}
    \sum X^+ + \sum X^- = 0, 
    \label{e3-1}
\end{equation}

\noindent where $X^+, X^- \in \mathbb{R}^D$ represent the positive and negative samples, respectively. Under Fisher's Linear Discriminant \citep{Fisher1936THEUO}, the optimal classifier $\Phi \in \mathbb{R}^D$ obtained by maximizing the ratio between the intra-class variance and the inter-class variance as: 
\begin{equation}
    \Phi = S_{\omega}^{-1} \cdot \left( \frac{1}{2N} \sum X^+ - \frac{1}{2N} \sum X^- \right) = S_{\omega}^{-1} \cdot \frac{1}{N} \sum X^+, 
    \label{e3-2}
\end{equation}

\noindent where $S_{\omega} = \Sigma^+ + \Sigma^- $, and $\Sigma$ means the sample covariance. The sample covariance of different categories is usually similar in a task. % \uncertain{ok}
% The binary classification task could be generalized to the multi-classification task by increasing the number of classifiers. 

\subsection{CAD with Fisher's Linear Discriminant}
\label{s3-2}

% $h_c$ and $h_r$ contain all the information implied by the corresponding text. 
Many studies on OOD generalization considered that features could be decomposed by their properties \citep{DBLP:conf/eccv/KhoslaZMET12,DBLP:conf/midl/IlseTLW20}. Therefore, we assume that each sentence consists of causal features $h_c \in \mathbb{R}^{D_c}$, whose joint distribution with labels is fixed for all environments, and correlated features $h_r \in \mathbb{R}^{D_r}$, whose distribution could vary, and the sentence could be sufficiently represented by $h_c$ and $h_r$. $D_c$ and $D_r$ denote the feature dimensions, respectively. Then, the classification of sentences $\{ S \}$ could be transformed into the classification of $\{ h_c \}$ and $\{ h_r \}$ as: 
\begin{equation}
    \Phi(S)=\Phi(h_c \oplus h_r). 
    \label{e3-3}
\end{equation}

We follow the assumptions about the feature distribution in \cite{Rosenfeld2021TheRO,Joshi2021AnIO}: in the binary classification setting, the labels $y \in \{-1, 1\}$ are drawn with equal probability, while both causal and correlated features could be drawn from a Gaussian distribution as: 
\begin{equation}
    h_c \sim \mathcal{N}(y \cdot \mu_c, \Sigma_c), \quad h_r \sim \mathcal{N}(y \cdot \mu_r, \Sigma_r), 
    \label{e3-4}
\end{equation}

\noindent where $\mu_c \in \mathbb{R}^{D_c}$; $\mu_r \in \mathbb{R}^{D_r}$; $\Sigma_c \in \mathbb{R}^{D_c \times D_c}$; $\Sigma_r \in \mathbb{R}^{D_r \times D_r}$. To simplify the subsequent analysis, we consider $\Sigma_c$ and $\Sigma_r$ as diagonal matrices. When Fisher's Linear Discriminant is employed, according to Eq. \ref{e3-2}, $S_{\omega} = 2(\Sigma_c \oplus \Sigma_r)$, and the optimal classifier for original dataset could be represented as: 
\begin{equation}
\begin{split}
    \Phi_{\text{ori}} &= \left(2(\Sigma_c \oplus \Sigma_r)\right)^{-1} \cdot 2(\mu_c \oplus \mu_r) \\
    & = \Sigma_c^{-1} \mu_c \oplus \Sigma_r^{-1} \mu_r = [\Phi_c , \Phi_r], 
\end{split}
    \label{e3-5}
\end{equation}

\begin{figure}[t]
    \centering
    \subfigure[Original Data]{
    \includegraphics[height=0.313\columnwidth]{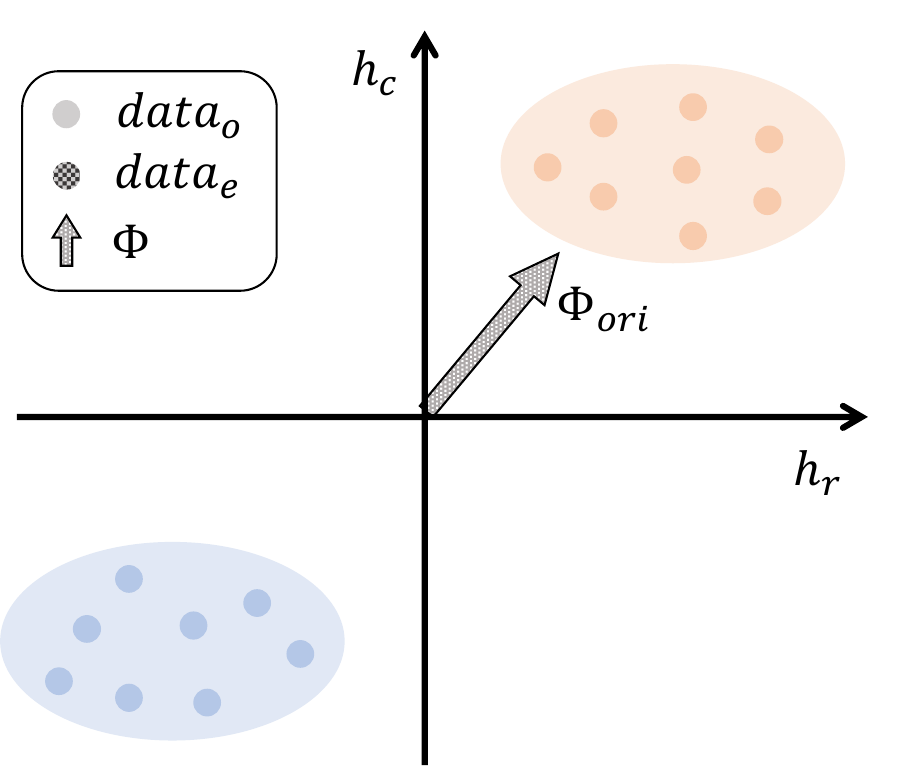} 
    \label{f3-1a}
    }
    \subfigure[CAD]{
    \includegraphics[height=0.313\columnwidth]{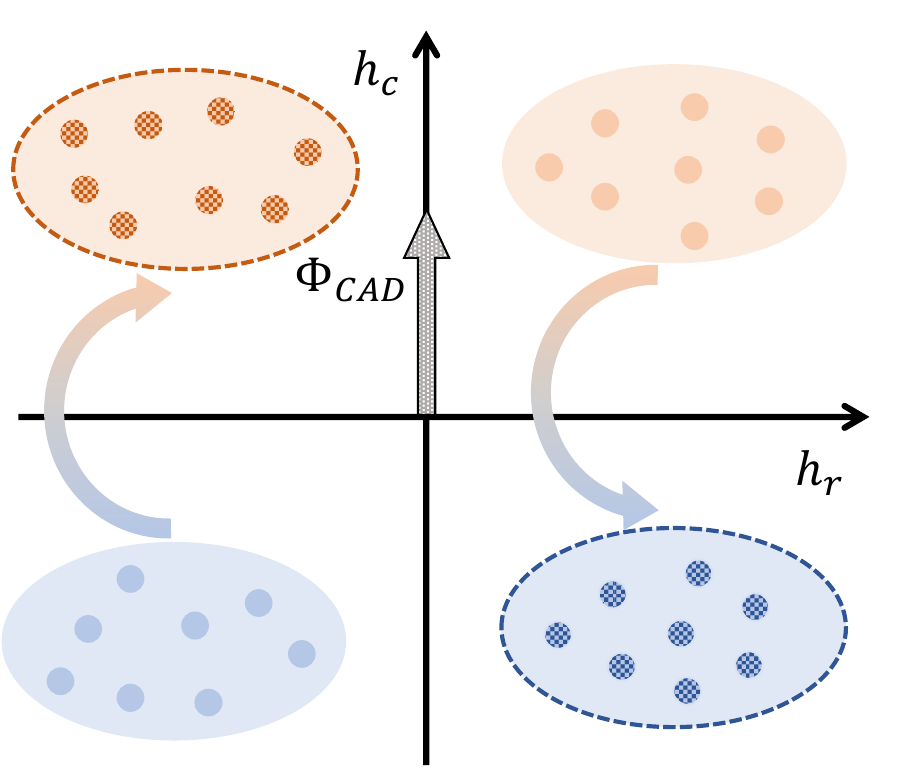} 
    \label{f3-1b}
    }
    \caption{Illustration of the effectiveness of CAD, where $data_o$ and $data_e$ denote the original and edited data features, respectively, and $\Phi$ denotes the optimal classifier. Compared to $\Phi_{\text{ori}}$ in Fig. \ref{f3-1a}, $\Phi_{\text{CAD}}$ in Fig. \ref{f3-1b} eliminates the interference of $h_r$ and is more robust in OOD generalization. }
    \label{f3-1}
\end{figure}
% The motivation of CAD. Counterfactual augmentation of texts (Fig. \ref{f1-1a}) changes the data distribution of the dataset (from Fig. \ref{f1-1b} to Fig. \ref{f1-1c}), which helps the model $\Phi$ to exploit causal features $h_c$ and exclude correlated features $h_r$. 

\noindent where $\Phi_c = \Sigma_c^{-1} \mu_c$ and $\Phi_r = \Sigma_r^{-1} \mu_r$ are the optimal classifiers in the causal and correlated feature spaces, respectively. Because the optimal classifier $\Phi_{\text{ori}}$ relies on the correlated features (Fig. \ref{f3-1a}), it may lead to performance degradation in OOD generalization. Ideally, a robust classifier in OOD generalization $\omega_{\text{rob}}$ should make predictions based on the causal features only as: 
\begin{equation}
    \Phi_{\text{rob}} = [\Phi_c, 0]. 
    \label{e3-6}
\end{equation}

CAD is introduced to mitigate language models' reliance on spurious correlations, and the principle of the augmentation operation is flipping the label with minimal editing. Therefore, the edited part of sentences corresponds to causal features $h_c$, while the correlated features $h_r$ are preserved, this is why CAD is expected to distinguish between correlated features and causal features. In the ideal situation, the causal features of each edited sentence are changed from $h_c$ to $-h_c$ by the counterfactual augmentation operation (Fig. \ref{f3-1b}). Following this change in the feature distribution, $\sum h^+_r - \sum h^-_r = 0$, while $\sum h^+_c - \sum h^-_c = 2 \mu_c$ is constant in CAD, and $S_{\omega} = 2(\Sigma_c \oplus (\mu_r \mu_r^T + \Sigma_r))$. Therefore, according to Eq. \ref{e3-2}, the optimal classifier for CAD is: 
\begin{equation}
\begin{split}
    \Phi_{\text{CAD}} &= \left(2(\Sigma_c \oplus (\mu_r \mu_r^T + \Sigma_r))\right)^{-1} \cdot 2(\mu_c \oplus 0) \\
    & = [\Phi_c, 0] = \Phi_{\text{rob}}. 
\end{split}
    \label{e3-7}
\end{equation}

\noindent The result in Eq. \ref{e3-7} illustrates the effectiveness of CAD. Intuitively, the counterfactual augmentation operation achieves a decoupling of $h_c$ and $h_r$, so the classifier is able to spontaneously exclude those dataset-specific spurious correlations. % \uncertain{ok}

\begin{figure*}[t]
    \centering
    % \vspace{-0.7cm}
    \includegraphics[width=0.95\columnwidth]{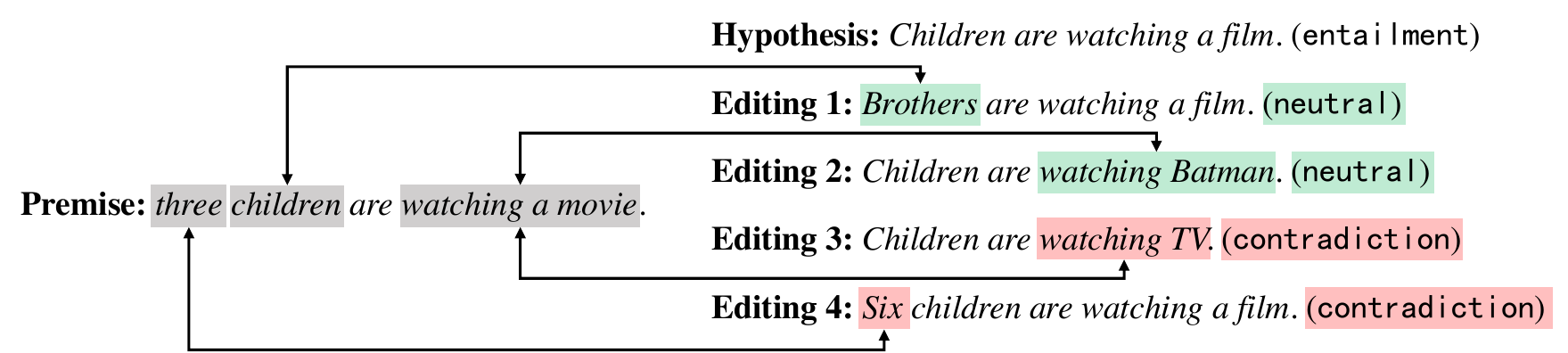}
    \caption{An example of multiple counterfactual augmentation operations in the Natural Language Inference task. Editing different causal components in the \textbf{Hypothesis} could all serve the purpose of flipping the corresponding label. }
    \label{f2-1}
    % \vspace{-0.5cm}
\end{figure*}

\section{Myopia Phenomenon in CAD}
\label{s4}

% Intuitively, by comparing the differences in counterfactual sentence pairs, language models could capture the features that have a causal effect on the labels. 
As mentioned in Section \ref{s3}, the essence of CAD is to change the data distribution through the counterfactual augmentation operation, thereby reducing the dataset-specific correlation bias implied in the data distribution. However, we find that the results of the counterfactual augmentation operation are diverse for a particular sentence, as illustrated in Fig. \ref{f2-1}. Specifically, the causal components and the perturbation types \citep{Joshi2021AnIO} (e.g., negation, quantifier, lexical, delete) that could flip labels are diverse, so different 
edited sentence could be obtained by making a specific perturbation for a particular causal component, while the other causal components remain unchanged. Therefore, compared to Eq. \ref{e1-1}, a more reasonable assumption is that only part of $h_c$ in counterfactual sentence pairs would change with the augmentation operation as: 
\begin{equation}
    \begin{split}
        \Phi(h_e, h_u, h_r) &= Y, \\
        \Phi(h_e^*, h_u, h_r) &= Y^*, 
    \end{split}
    \label{e4-1}
\end{equation}

\noindent where $h_c$ is distinguished into the edited features $h_e$ that would change with the augmentation operation and the non-edited features $h_u$ that would not change. This assumption is empirically convincing because of the analysis and experiments in \cite{Joshi2021AnIO,Kaushik2020LearningTD}. Similar to the analysis of Eq. \ref{e1-1}, Eq. \ref{e4-1} gives us an important insight: language models trained on original data and CAD focus on different features in counterfactual sentence pairs. On the one hand, CAD eliminates the interference of correlated features; on the other hand, language models inevitably exclude non-edited causal features. In this study, we refer to this as the myopia phenomenon. % \uncertain{ok}

\subsection*{Myopia Phenomenon Under Fisher's Linear Discriminant}

% \citep{Balashankar2021CanWI} emphasized the importance of all causal features to improve model robustness. Therefore, the causal features $h_c$ are divided into edited features $h_e$ and non-edited features $h_u$, and better augmentation operations correspond to the less $h_u$. 
Here, we analyze the myopia phenomenon from the perspective of Fisher's Linear Discriminant. Since the augmentation operation could only change part of the causal features, the edited features $h_e$ and the non-edited features $h_u$ are independent of each other. Therefore, without loss of generality, we assume that the distributions of $h_e$ and $h_u$ as: 
\begin{equation}
    h_e \sim \mathcal{N}(y \cdot \mu_e, \Sigma_e), \quad h_u \sim \mathcal{N}(y \cdot \mu_u, \Sigma_u), 
    \label{e4-2}
\end{equation}

\noindent where $\mu_c = \mu_e \oplus \mu_u $; $\Sigma_c = \Sigma_e \oplus \Sigma_u $. Under this more reasonable setting, we follow the method in Section \ref{s3-2} to employ Fisher's Linear Discriminant to obtain three optimal classifiers as:
\begin{equation}
    \begin{split}
        \Phi_{\text{rob}} &= [\Phi_e, \Phi_u, 0], \\
        \Phi_{\text{ori}} &= [\Phi_e, \Phi_u, \Phi_r], \\
        \Phi_{\text{CAD}} &= [\Phi_e, 0, 0], \\
    \end{split}
    \label{e4-3}
\end{equation}

\noindent where $\Phi_e = \Sigma_e^{-1} \mu_e$ and $\Phi_u = \Sigma_u^{-1} \mu_u$ are the optimal classifiers in the edited and non-edited feature spaces, respectively. It should be noted that $\Phi_c = [\Phi_e, \Phi_u]$. Here, we employ cosine similarity \citep{Park2020AMC} as the metric of similarity to the robust classifier $\Phi_{\text{rob}}$, and the similarity of $\Phi_{\text{ori}}$ and $\Phi_{\text{CAD}}$ could be denoted as: 
\begin{equation}
    \begin{split}
        \cos{\theta_{\text{ori}}} &= \frac{1}{\sqrt{1+\frac{|\Phi_r|^2}{|\Phi_e|^2 + |\Phi_u|^2}}}, \\
        \cos{\theta_{\text{CAD}}} &= \frac{1}{\sqrt{1+\frac{|\Phi_u|^2}{|\Phi_e|^2}}}. \\
    \end{split}
    \label{e4-4}
\end{equation}

\noindent Eq. \ref{e4-4} quantifies the myopia phenomenon of CAD through Fisher's Linear Discriminant. Specifically, the similarity of $\Phi_{\text{ori}}$ depends on the ratio of $|\Phi_r|$ to $|\Phi_c|$, while the similarity of $\Phi_{\text{CAD}}$ depends on the ratio of $|\Phi_u|$ to $|\Phi_e|$. Therefore, it is uncertain whether $\Phi_{\text{ori}}$ or $\Phi_{\text{CAD}}$ is better or worse. Intuitively, for simple tasks (e.g. Sentiment Analysis), since there are relatively few causal features, the augmentation operation is prone to focus on most of the causal features, so CAD tends to exhibit great OOD generalization capability; while complex tasks (e.g. Natural Language Inference) have more causal features, so the improvement in OOD generalization capability is not significant, as a large number of causal features are not included by the augmentation operation. This analysis is consistent with existing studies \citep{Huang2020CounterfactuallyAugmentedST,Joshi2021AnIO}. % \uncertain{ok}

\section{Proposed Method}
\label{s5}

To address the myopia phenomenon in Section \ref{s4}, we require to \textbf{E}xtract more complete \textbf{C}ausal \textbf{F}eatures as well as exclude correlated features. In this section, we propose two insights on the structural properties of CAD at the dataset level and the sentence level, and then design additional constraints based on these insights, to further unlock the generalization potential of CAD. In this study, we refer to this method as the ECF algorithm. % \uncertain{ok}

\subsection{Dataset-Level Constraint}

\emph{\textbf{Insight:} the data distribution of the original data could alleviate the myopia phenomenon of CAD. }

% Previous experiments \citep{Kaushik2020LearningTD} have also revealed that models trained on original data performed poorly on corresponding CAD and vice versa. 
Due to the change in data distribution, the features that language models focus on are different: language models with CAD only focus on the edited causal features $h_e$ (myopia phenomenon), while language models with the original data confuse $h_c$ and $h_r$ (but no myopia phenomenon). Different data distributions lead to different situations, which indicates that the original data distribution carries information that is missing in CAD. Therefore, there are potential complementary effects of the original data and CAD on causal feature extraction. 

In fact, this complementary could also be observed in the case of Fisher's Linear Discriminant. Specifically, we find that the two classifiers $\Phi_{\text{ori}}$ and $\Phi_{\text{CAD}}$ in Eq. \ref{e4-3} are complementary. In the simplest case, the interpolated classifier $\Phi_{\text{I}} = \lambda \cdot \Phi_{\text{ori}} + (1 - \lambda) \cdot \Phi_{\text{CAD}}$ could achieve a better cosine similarity than both $\Phi_{\text{ori}}$ and $\Phi_{\text{CAD}}$ when $\lambda \in (0, 1)$. When $\lambda = \frac{|\Phi_u|^2}{|\Phi_u|^2 + |\Phi_r|^2}$, the cosine similarity between $\Phi_{\text{I}}$ and $\Phi_{\text{rob}}$ is maximized as: 
\begin{equation}
    \begin{split}
        \cos{\theta_{\text{I}}} &= \frac{|\Phi_e|^2+\frac{|\Phi_u|^4}{|\Phi_u|^2+|\Phi_r|^2}}{\sqrt{|\Phi_e|^2+|\Phi_u|^2}\sqrt{|\Phi_e|^2+\frac{|\Phi_u|^6+|\Phi_u|^4 |\Phi_r|^2}{(|\Phi_u|^2 + |\Phi_r|^2)^2}}} \\
        &> \max \{\cos{\theta_{\text{ori}}}, \cos{\theta_{\text{CAD}}} \}. 
    \end{split}
    \label{(e5-1}
\end{equation}

\noindent Therefore, we consider that it is possible to exploit the complementary nature of the two data distributions to unlock the OOD generalization potential of CAD. 

% According to the analysis in Section \ref{s2}, the original data and CAD have different feature distributions, 
% so we consider them as two training environments $\mathcal{E}_{tr} = \{ e_{\text{ori}}, e_{\text{CAD}} \}$, and use IRM to fuse the advantages of both. 
% Although CAD is obtained by extending the original data, this extension is non-identically distributed and could change the feature distribution in the dataset. 
In this study, we adopt the Invariant Risk Minimization (IRM) method \citep{DBLP:journals/corr/abs-1907-02893,Rosenfeld2021TheRO} to exploit this complementary nature. The role of IRM is to estimate the invariant causal features from multiple training environments. As mentioned in Section \ref{s3}, the counterfactual augmentation operation does not follow the IID principle, which allows us to consider the original data and CAD as two different training environments $\mathcal{E}_{tr} = \{ \mathcal{E}_{\text{ori}}, \mathcal{E}_{\text{CAD}} \}$, and then it is reasonable to adopt the IRM method to fuse the advantages of both environments. Specifically, to induce the language model $M$ to learn the invariant causal features across environments, the additional constraint $\mathcal{L}_{IRM}$ is designed as: 
\begin{equation}
    \mathcal{L}_{IRM} = \sum_{\mathcal{E} \in \mathcal{E}_{tr}} \Vert \nabla_{\omega | \omega=1.0} \mathcal{R}_{\mathcal{E}}(\omega \cdot M) \Vert^2, 
    \label{(e5-2}
\end{equation}

% To learn invariances across environments, find a data representation such that the optimal classifier on top of that representation matches all environments. 
\noindent where $\mathcal{R}_{\mathcal{E}}(\cdot)$ is the prediction risk loss under environment $\mathcal{E}$, and $\omega = 1.0$ as a scalar is a fixed ``dummy’' classifier. The essence of $\mathcal{L}_{IRM}$ \citep{DBLP:journals/corr/abs-1907-02893} is a gradient norm penalty that measures the optimality of the ``dummy’' classifier in each environment, in order to find the invariant causal features that match all environments. % \uncertain{ok}

\subsection{Sentence-Level Constraint}

\emph{\textbf{Insight:} the correlated features $h_r$ of counterfactual sentence pairs are not guaranteed to be aligned. }

% it is consistent with the original design of CAD to make an explicit constraint on $h_r$ for counterfactual sentence pairs. 
In our assumption for CAD in Section \ref{s4}, the correlated features $h_r$ of counterfactual sentence pairs are consistent, as we consider that the counterfactual augmentation operation would only affect part of $h_c$. However, this property is not guaranteed for language models trained directly on CAD, and this potential dissimilarity of $h_r$ gives language models the convenience to utilize $h_r$. In Fisher's Linear Discriminant perspective, the potential dissimilarity of $h_r$ could lead to $\sum h^+_r - \sum h^-_r = \Delta h_r \neq 0$, and the component of the correlated feature space in current $\Phi_{\text{CAD}}$ would become: 
\begin{equation}
    \Phi_r = \left(2(\mu_r \mu_r^T + \Sigma_r)\right)^{-1} \cdot \frac{1}{2N}(\Delta h_r) \neq 0, 
    \label{(e5-3}
\end{equation}

\noindent which is detrimental to the robustness of $\Phi_{\text{CAD}}$ because it would exploit the correlation features. Therefore, it is reasonable to design an explicit constraint to align $h_r$ of counterfactual sentence pairs. 

\begin{figure}[t]
    \centering
    \includegraphics[height=0.313\columnwidth]{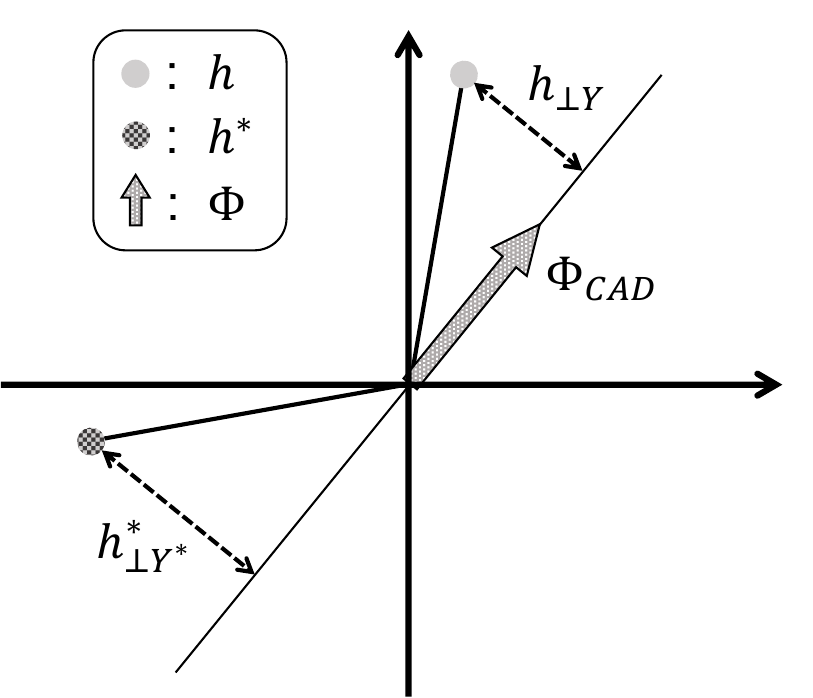}
    \caption{Illustration of sentence feature decomposition. In binary classification, for a counterfactual sentence feature pair $(h, h^*)$, the orthogonal components $(h_{\perp Y}, h^*_{\perp Y^*})$ along the feature classifier $\Phi$ are considered as proxies for the correlated features. }
    \label{f5-1}
\end{figure}

% most models could be divided into a feature extraction module $\Psi$ and a Fully-Connected layer (concatenated by label vectors $h_Y$
However, $h_r$ and $h_c$ in CAD are difficult to decompose in language models, so we should find a sensible proxy for $h_r$. Noting that $h_r$ has little effect on the prediction in CAD, based on this property, we creatively construct the proxy of $h_r$ through the mechanism of the feature classifier. Most feature classifiers are fully-connected layers, where each row of the weight matrix could be interpreted as a label vector $h_Y$ \citep{Du2019ExplicitIM}, and the label probability could be obtained by the dot product of the sentence vector $h$ and each label vector $h_Y$ as: 
\begin{equation}
    p(y_k) = \frac{\text{exp}(h_Y^k \cdot h)}{\sum_{i=1}^N\text{exp}(h_Y^i \cdot h)}. 
    \label{(e5-4}
\end{equation}

\noindent In this way, $h$ could be decomposed along $h_Y$, where the parallel component $h_{\parallel Y}$ determines the label probability and the orthogonal component $h_{\perp Y}$ has no effect on the prediction. The commonality between $h_{\perp Y}$ and $h_r$ makes $h_{\perp Y}$ an ideal proxy for $h_r$, as shown in Fig. \ref{f5-1}. Specifically, for a counterfactual sentence feature pair $(h, h^*)$, we design $\mathcal{L}_{OCD}$ to penalize their Orthogonal Component Distance as: 
\begin{equation}
    \mathcal{L}_{OCD} = \Vert h_{\perp Y} - h^*_{\perp Y^*} \Vert^2. 
    \label{(e5-5}
\end{equation}

\noindent This is a positive feedback process, so even if initially the feature classifier has a large estimation error, it would gradually become accurate with the help of the prediction loss and $\mathcal{L}_{OCD}$. % \uncertain{ok}

\subsection{Training Process}

\begin{figure}[t]
    \centering
    \includegraphics[width=0.8\columnwidth]{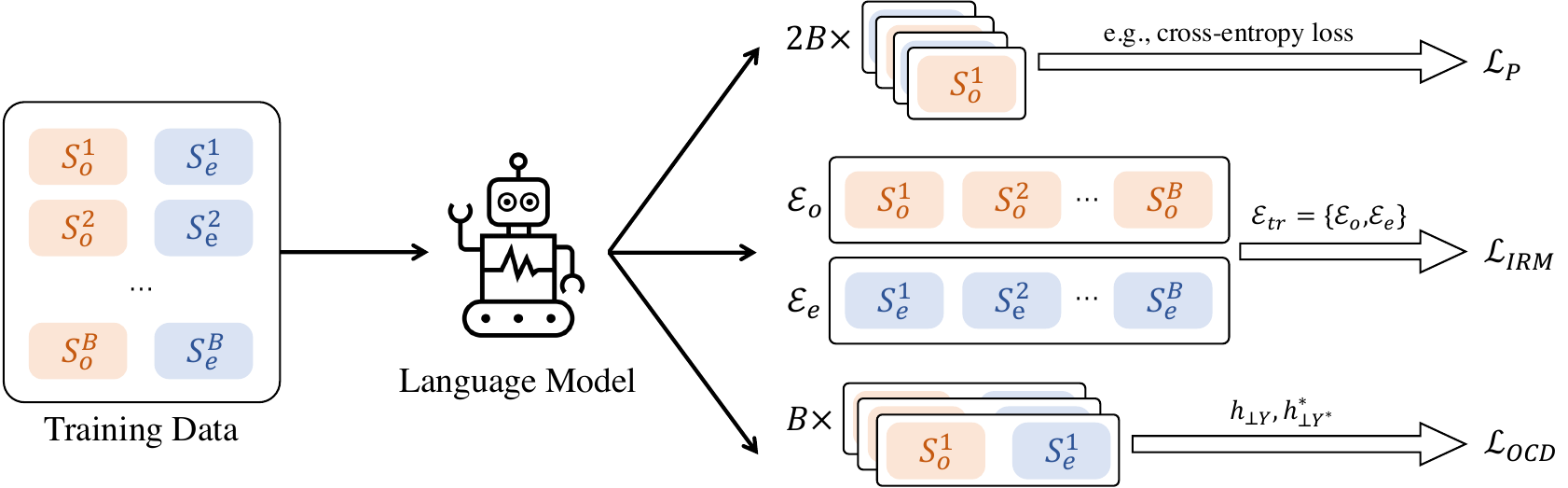}
    \caption{The optimization process of the ECF algorithm. }
    \label{f5-2}
\end{figure}

Compared to the traditional prediction loss $\mathcal{L}_{P}$ (e.g., cross-entropy loss), the ECF algorithm also combines dataset-level constraint $\mathcal{L}_{IRM}$ and sentence-level constraint $\mathcal{L}_{OCD}$, and the optimization process of the ECF algorithm is shown in Fig. \ref{f5-2}. Specifically, during the training process, each batch contains $B$ counterfactual sentence pairs $\{S_o, S_e\}_B$, and the three loss terms are optimized for different objects: $\mathcal{L}_{IRM}$ treats $\{S_o\}_N$ and $\{S_e\}_N$ as training samples from two different environments; $\mathcal{L}_{OCD}$ calculates the loss for each sentence pair $\{S_o, S_e\}$ as a minimum unit; $\mathcal{L}_{P}$ directly optimizes the prediction loss of each sentence. Therefore, the loss of the ECF algorithm is denoted as: 
\begin{equation}
    \mathcal{L} = \mathcal{L}_{P} + \alpha \cdot \mathcal{L}_{IRM} + \beta \cdot \mathcal{L}_{OCD}, 
    \label{(e5-6}
\end{equation}

\noindent where $\alpha$, $\beta$ are the weighting coefficients to balance the language model's In-Distribution predictive capability and additional constraints introduced for OOD generalization. % \uncertain{ok}

\section{Experiments}
\label{s6}

\subsection{Datasets}

We conducted experiments on two tasks: Sentiment Analysis (SA) and Natural Language Inference (NLI). Generally, CAD is obtained by extending a seed dataset with counterfactual augmentation operations. % \uncertain{ok}

\noindent \textbf{Sentiment Analysis} \quad The seed dataset in SA was a subset of IMDb \citep{Maas2011LearningWV} dataset. \cite{Kaushik2020LearningTD} manually annotated the corresponding counterfactual sentences in the seed dataset to construct $\text{CAD}_{h}$, while \cite{Yang2021ExploringTE,Wang2021RobustnessTS} utilized WordNet \citep{Fellbaum2000WordNetA} to automatically generate counterfactual sentences and constructed $\text{CAD}_{a}$. We evaluated each language model's OOD generalization capability on three OOD datasets: SST-2 \citep{Socher2013RecursiveDM}, Amazon review \citep{Ni2019JustifyingRU}, Yelp review \citep{Zhang2015CharacterlevelCN}. % \uncertain{ok}

\noindent \textbf{Natural Language Inference} \quad The seed dataset in NLI was a subset of SNLI \citep{Bowman2015ALA} dataset, and \cite{Kaushik2020LearningTD} constructed $\text{CAD}_{h}$ by manually editing the seed dataset. Because the NLI task is more complex, there is little research related to the automatic generation of counterfactual sentences. We employed MNLI (split into matched and mismatched parts) \citep{Williams2018ABC} as our OOD dataset for evaluation of language models. % \uncertain{ok}

\begin{table*}[t]
  \centering
%   \fontsize{10}{10}\selectfont
  \begin{threeparttable}
    \resizebox{12.1cm}{!}{
    \begin{tabular}{lccccccc}
    \toprule
    \bf Model & \bf Task & \bf Embedding Dimension & \bf Batch Size & \bf Learning Rate & \bf Epochs & \bf $\alpha$ & \bf $\beta$ \cr
    \midrule
    BiLSTM & SA & 300 & 32 & 1e-3 & 100 & 1.6 & 0.1 \cr
    BERT & SA & 768 & 8 & 1e-5 & 10 & 0.1 & 0.1 \cr
    Roberta & SA & 768 & 8 & 1e-5 & 10 & 0.1 & 0.1 \cr
    BiLSTM & NLI & 300 & 30 & 1e-3 & 100 & 1.6 & 0.1 \cr
    BERT & NLI & 768 & 5 & 1e-5 & 10 & 0.1 & 0.1 \cr
    Roberta & NLI & 768 & 5 & 1e-5 & 10 & 0.1 & 0.1 \cr
    \bottomrule
    \end{tabular}
    }
    \end{threeparttable}
    \caption{Hyperparameter setting of the experiments. }
    \label{taba-1}
\end{table*}

\subsection{Implementation Details}

In our experiments, we implemented BiLSTM \citep{Hochreiter1997LongSM} and pre-trained models BERT \citep{Devlin2019BERTPO}, Roberta \citep{Liu2019RoBERTaAR} as the base models, and employed the Adam optimizer to update all network parameters. We selected the best checkpoint on the training set for evaluation. For pre-trained models, we used the Hugging Face implementation \citep{Wolf2019HuggingFacesTS} to fine-tune the pre-trained models. The hyperparameters of the experiments are listed in Table \ref{taba-1}. We selected hyperparameters $\alpha$ and $\beta$ based on the guideline that multiple losses should be weighted to the same order of magnitude, which facilitated language models to take full advantage of additional constraints and were not dominated by a single loss term. In our experiments, datasets did not show strong sensitivity to $\alpha$ and $\beta$. We consider this is because the losses $\mathcal{L}_{IRM}$ and $\mathcal{L}_{OCD}$ corresponding to $\alpha$ and $\beta$ are both committed to improving the capability of language models to extract complete causal features $h_c$, so there is no trade-off relationship between them. % \uncertain{ok}
% However, pre-trained models and BiLSTM corresponded to different hyper-parameters (still following the guideline to select hyper-parameters), and we speculated that different models mapping texts to different feature spaces led to this situation. 

\begin{table}[!t]
  \centering
%   \fontsize{10}{10}\selectfont
  \begin{threeparttable}
    \resizebox{12.1cm}{!}{
    \begin{tabular}{llcccccc}
    \toprule
    \bf Dataset & \bf Method & \bf Original & \bf CAD & \bf SST-2 & \bf Amazon & \bf Yelp & \bf Mean \cr
    \midrule
    \multirow{3}{*}{Seed} & BiLSTM & 74.3 & 64.0 & 64.5 & 61.7 & 62.3 & 62.8 \cr
     & BERT & 85.2 & 85.5 & 82.8 & 88.0 & 85.5 & 85.4 \cr
     & Roberta & 88.0 & 89.2 & 86.4 & 91.5 & 93.0 & 90.3 \cr
    \midrule
     & BiLSTM & 80.5 & 85.0 & 65.6 & 71.6 & 74.7 & 70.6 \cr
     & BiLSTM+ECF & 80.7 & 84.4 & \bf 71.4 & \bf 74.0 & \bf 77.4 & \bf 74.3 \cr
    \cmidrule(lr){2-8}
    $\text{CAD}_{h}$ & BERT & 84.7 & 88.8 & 84.7 & 90.0 & 90.6 & 88.4 \cr
    \citep{Kaushik2020LearningTD} & BERT+ECF & 84.7 & 89.2 & \bf 84.9 & \bf 92.9 & \bf 92.4 & \bf 90.1 \cr
    \cmidrule(lr){2-8}
     & Roberta & 87.5 & 92.3 & 85.5 & 94.3 & \bf 94.9 & 91.6 \cr
     & Roberta+ECF & 90.3 & 93.0 & \bf 89.6 & \bf 94.6 & 94.8 & \bf 93.0 \cr
    \midrule
     & BiLSTM & 56.1 & 66.7 & 61.5 & 57.6 & 57.9 & 59.0 \cr
     & BiLSTM+ECF & 57.9 & 67.5 & \bf 62.4 & \bf 59.0 & \bf 58.5 & \bf 60.0 \cr
    \cmidrule(lr){2-8}
    $\text{CAD}_{a}$ & BERT & 55.1 & 72.1 & 75.9 & 84.7 & 83.1 & 81.2 \cr
    \citep{Yang2021ExploringTE} & BERT+ECF & 80.4 & 63.8 & \bf 78.9 & \bf 88.1 & \bf 86.2 & \bf 84.4 \cr
    \cmidrule(lr){2-8}
     & Roberta & 61.7 & 70.5 & 74.7 & 85.2 & 84.8 & 81.6 \cr
     & Roberta+ECF & 61.9 & 69.5 & \bf 83.2 & \bf 90.1 & \bf 89.3 & \bf 87.5 \cr
    \midrule
     & BiLSTM & 67.8 & 75.4 & 59.8 & 63.3 & 62.6 & 61.9 \cr
     & BiLSTM+ECF & 76.1 & 79.7 & \bf 61.7 & \bf 66.4 & \bf 63.8 & \bf 64.0 \cr
    \cmidrule(lr){2-8}
    $\text{CAD}_{a}$ & BERT & 87.1 & 88.0 & 83.2 & 88.4 & 88.9 & 86.8 \cr
    \citep{Wang2021RobustnessTS} & BERT+ECF & 85.7 & 77.8 & \bf 83.6 & \bf 90.3 & \bf 89.5 & \bf 87.8 \cr
    \cmidrule(lr){2-8}
     & Roberta & 90.2 & 82.8 & 74.8 & 89.8 & 86.9 & 83.8 \cr
     & Roberta+ECF & 92.2 & 78.7 & \bf 84.5 & \bf 91.7 & \bf 90.3 & \bf 88.8 \cr
    \bottomrule
    \end{tabular}
    }
    \end{threeparttable}
    % \vspace{-0.1cm}
    \caption{Accuracy of different language models and datasets in SA. The best performance is \textbf{bold}. $\text{CAD}_{h}$ and $\text{CAD}_{a}$ represent manually annotated CAD and automatically generated CAD, respectively. }
    \label{tab5-1}
    % \vspace{-0.5cm}
\end{table}

\section{Results and Analysis}
\label{s7}

% This section is mainly about the performance of the ECF algorithm on OOD generalization. We reported the experimental results of the ECF algorithm over base models on two tasks (Section \ref{s7-1}). In addition, we verified the contribution of the two constraints in the ECF algorithm separately through the ablation experiments (Section \ref{s7-2}). To demonstrate the effectiveness of CAD for OOD generalization, we compared the data efficiency of CAD with the original data under different tasks and different language models (Section \ref{s7-3}). We also identified the myopia phenomenon caused by CAD through the case analysis and confirmed that the ECF algorithm could effectively alleviate the myopia phenomenon (Section \ref{s7-4}). % \uncertain{ok}

\subsection{Main Results}
\label{s7-1}

\noindent \textbf{Results on SA} \quad The results are displayed in Table \ref{tab5-1}. We found that the ECF algorithm helped all the base models to improve their accuracy on OOD datasets, which demonstrated the effectiveness of the ECF algorithm. Specifically, experiments were carried out on a manually annotated CAD and two automatically generated CADs. $\text{CAD}_{h}$ was more effective for language models' generalization, and the ECF algorithm improved the average accuracy of BiLSTM, BERT and Roberta on OOD datasets by 3.7\%, 1.7\% and 1.4\%, respectively. While, the language models trained on $\text{CAD}_{a}$ were relatively weaker in OOD generalization, and the ECF algorithm also led to an improvement in the average accuracy of BiLSTM, BERT and Roberta by 1.0\% / 3.2\% / 5.9\% and 2.1\% / 1.0\% / 5.0\% on two $\text{CAD}_{a}$, respectively. % \uncertain{ok}

% % \uncertain{?} and it was worth mentioning that the generalization capability of BiLSTM for $\text{CAD}_{h}$ was even worse than that for the seed dataset. 
\noindent \textbf{Results on NLI} \quad The results are displayed in Table \ref{tab5-2}. The results showed that the effectiveness of the ECF algorithm continued to hold on the NLI task. Specifically, the ECF algorithm improved the average accuracy of BiLSTM on OOD datasets by 1.5\%. At the same time, the ECF algorithm also helped pre-trained models, resulting in a 1.2\% average accuracy improvement for BERT and a 1.7\% average accuracy improvement for Roberta. % \uncertain{ok}

\begin{table}[t]
  \centering
%   \fontsize{10}{10}\selectfont
  \begin{threeparttable}
    \resizebox{12.1cm}{!}{
    \begin{tabular}{llccccc}
    \toprule
    \bf Dataset & \bf Method & \bf Original & \bf CAD & \bf MNLI-m & \bf MNLI-mm & \bf Mean \cr
    \midrule
    \multirow{3}{*}{Seed} & BiLSTM & 41.8 & 33.9 & 35.9 & 35.0 & 35.5 \cr
     & BERT & 71.5 & 53.8 & 53.6 & 55.1 & 54.3 \cr
     & Roberta & 83.8 & 67.2 & 67.4 & 68.4 & 67.9 \cr
    \midrule
     & BiLSTM & 39.8 & 39.0 & 34.4 & 35.0 & 34.7 \cr
     & BiLSTM+ECF & 44.2 & 37.6 & \bf 36.2 & \bf 36.2 & \bf 36.2 \cr
    \cmidrule(lr){2-7}
    $\text{CAD}_{h}$ & BERT & 79.2 & 71.2 & 62.6 & 64.3 & 63.5 \cr
    \citep{Kaushik2020LearningTD} & BERT+ECF & 77.0 & 72.0 & \bf 64.0 & \bf 65.5 & \bf 64.7 \cr
    \cmidrule(lr){2-7}
     & Roberta & 80.2 & 75.4 & 70.5 & 71.5 & 71.0 \cr
     & Roberta+ECF & 82.5 & 76.7 & \bf 72.6 & \bf 72.8 & \bf 72.7 \cr
    \bottomrule
    \end{tabular}
    }
    \end{threeparttable}
    % \vspace{-0.1cm}
    \caption{Accuracy of different language models in NLI. The best performance is \textbf{bold}. }
    \label{tab5-2}
    % \vspace{-0.3cm}
\end{table}

\subsection{Ablation Study}
\label{s7-2}

\begin{figure}[t]
    \centering
    \subfigure[BiLSTM]{
    \includegraphics[height=0.313\columnwidth]{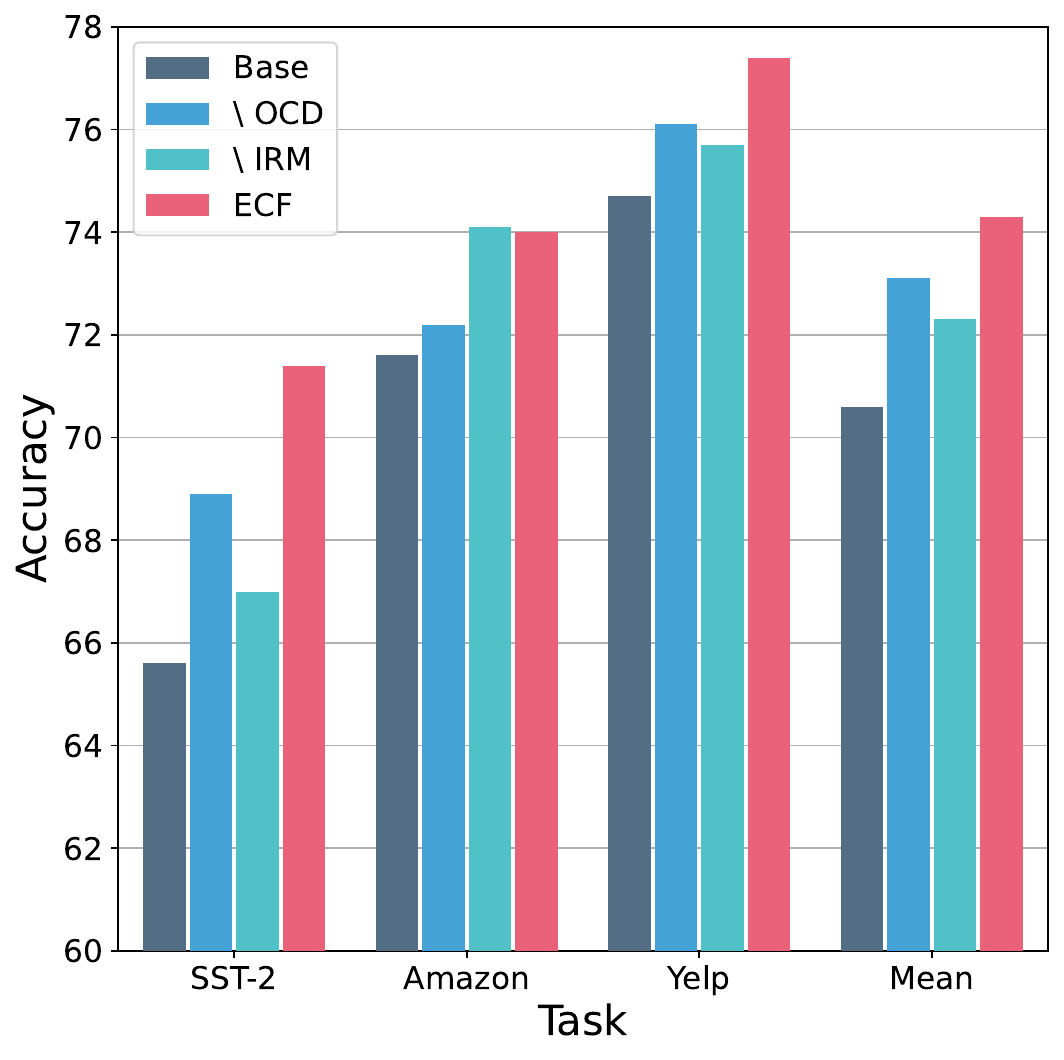} 
    \label{f6-1a}
    }
    \subfigure[BERT]{
    \includegraphics[height=0.313\columnwidth]{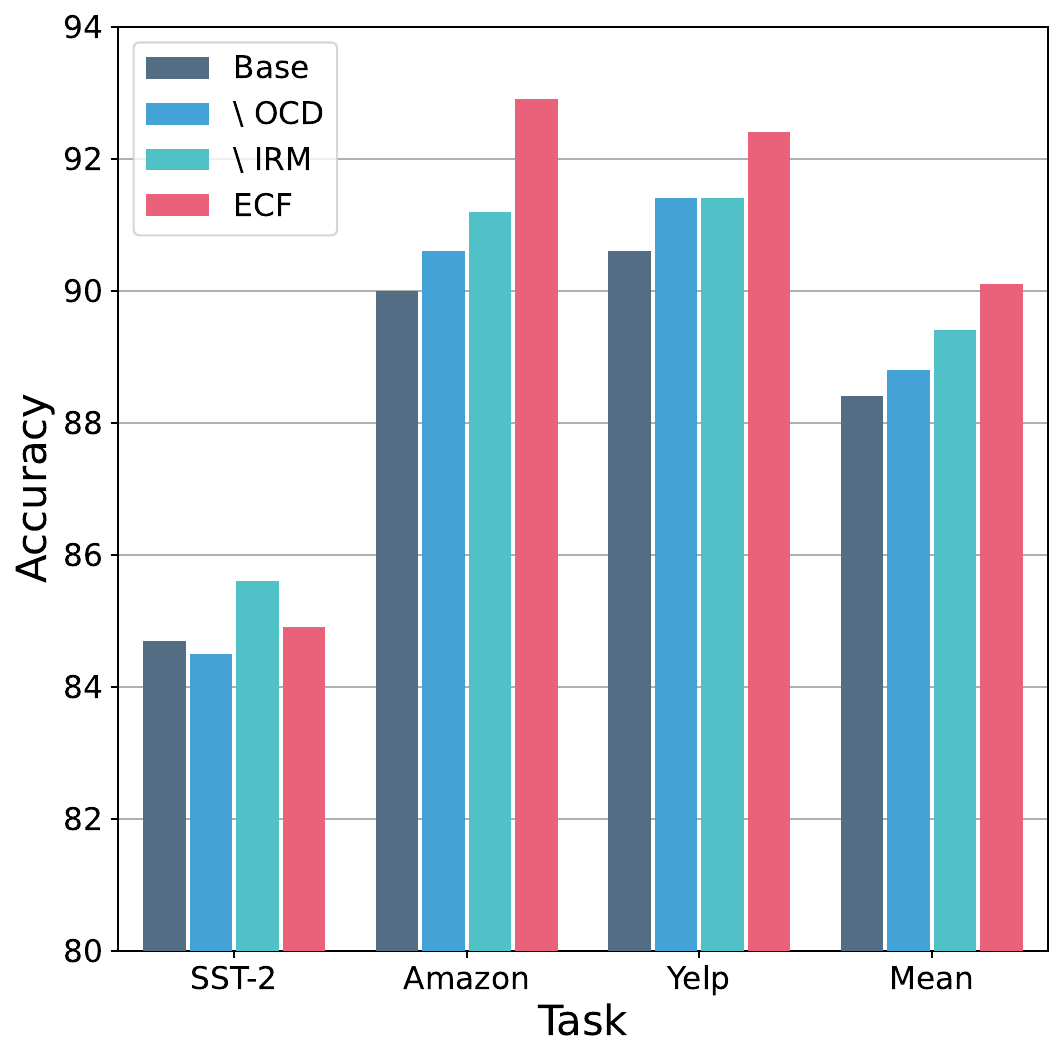} 
    \label{f6-1b}
    }
    \subfigure[Roberta]{
    \includegraphics[height=0.313\columnwidth]{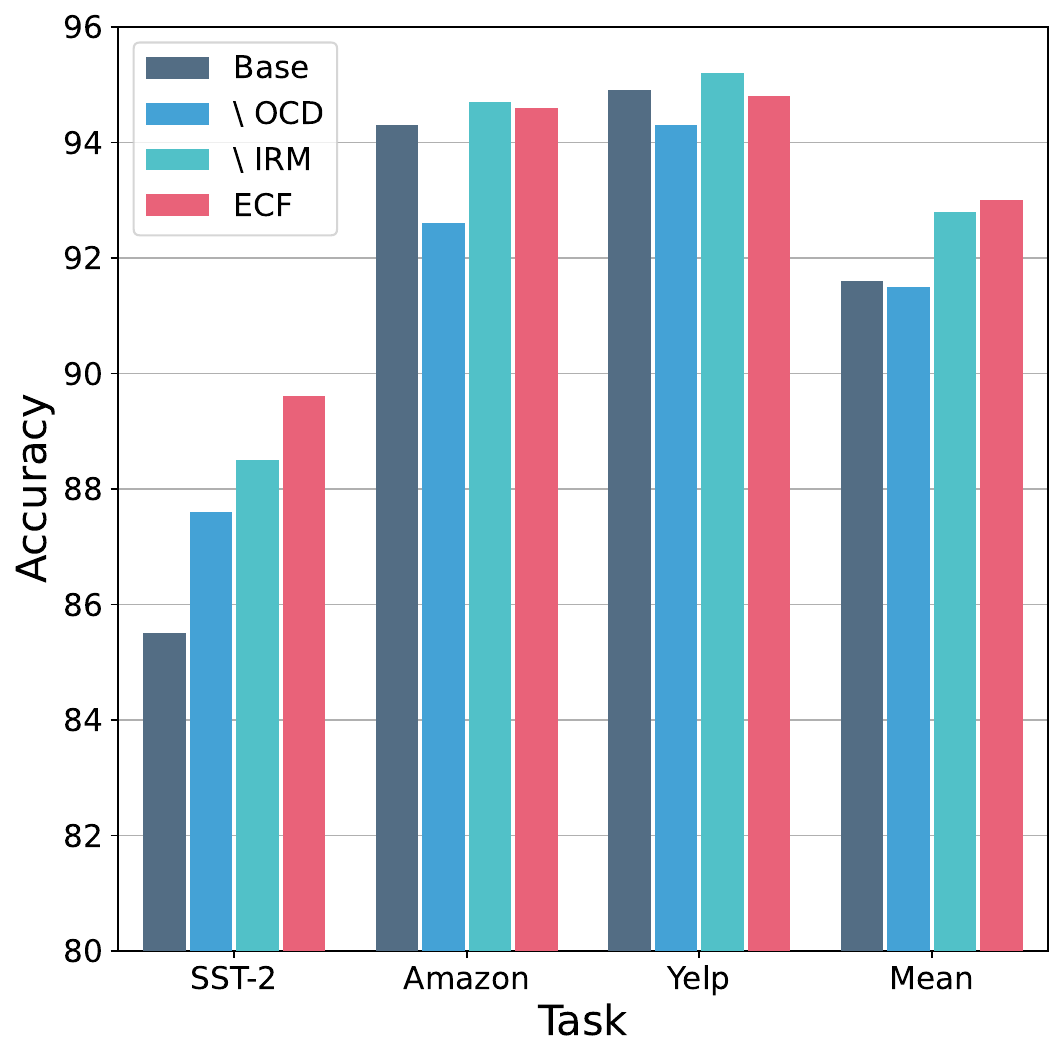} 
    \label{f6-1c}
    }
    \caption{Ablation analysis of two constraints on Sentiment Analysis. $\backslash$ denotes the removing operation. }
    \label{f6-1}
\end{figure}

We investigated the independent impact of each constraint in the ECF algorithm. We chose $\text{CAD}_{h}$ in the SA task as the dataset and conducted ablation experiments on BiLSTM, BERT and Roberta. The results are reported in Fig. \ref{f6-1}. Specifically, in most cases, the generalization capability of language models decreased significantly when the constraints $\mathcal{L}_{IRM}$ and $\mathcal{L}_{OCD}$ in the ECF algorithm were removed. This result is in line with our expectations: the two additional constraints based on CAD's properties could unlock CAD's potential for OOD generalization. % \uncertain{ok}

\subsection{Data Efficiency}
\label{s7-3}

\begin{figure}[t]
    \centering
    \subfigure[SA]{
    \includegraphics[height=0.313\columnwidth]{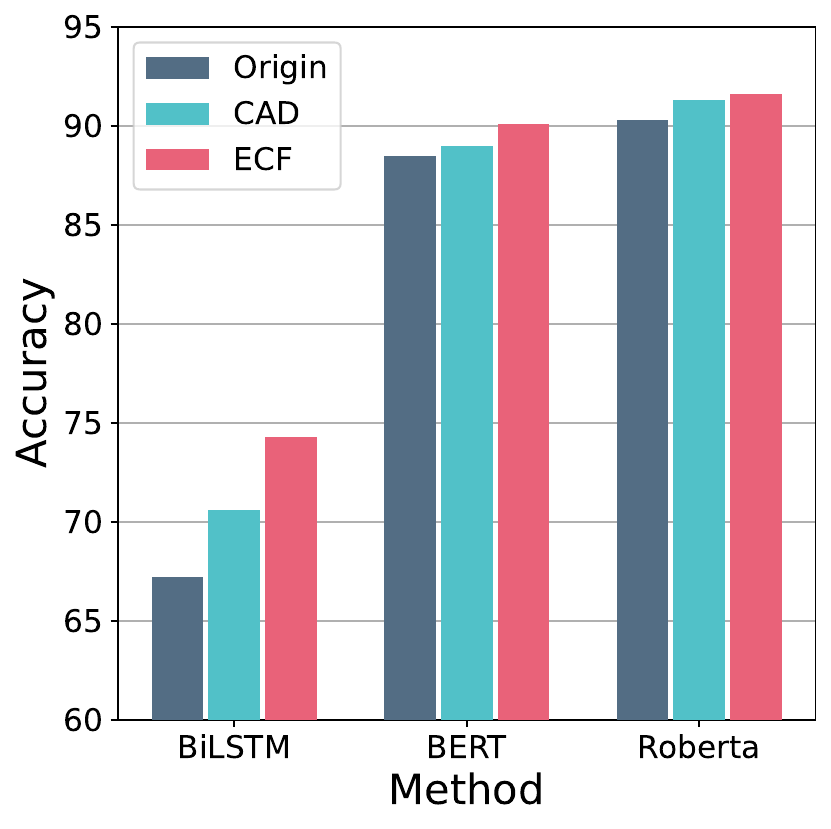} 
    \label{f6-2a}
    }
    \subfigure[NLI]{
    \includegraphics[height=0.313\columnwidth]{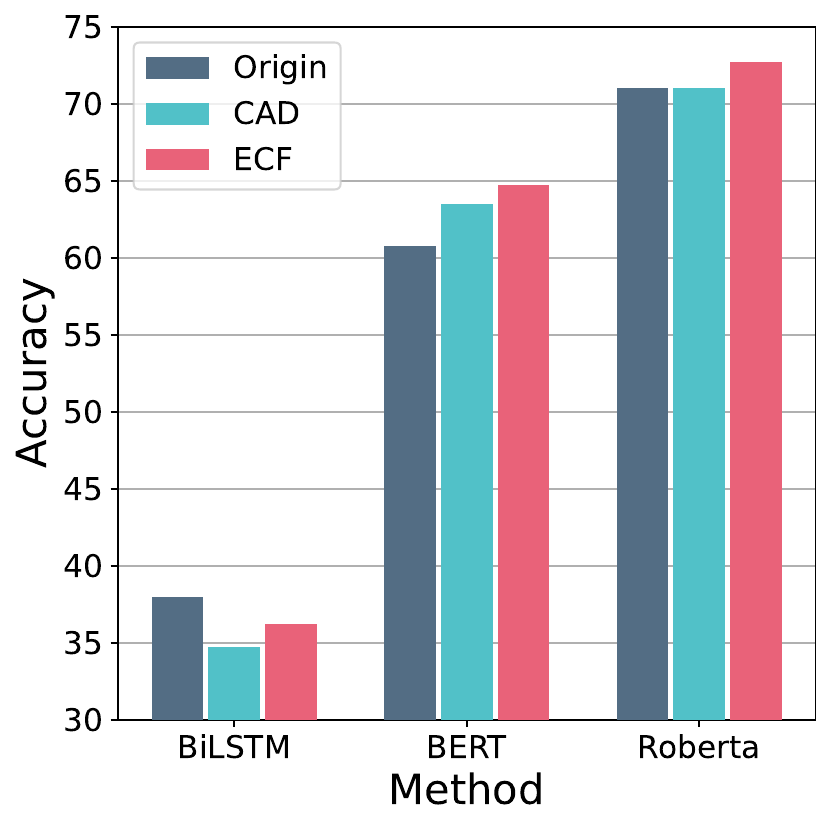} 
    \label{f6-2b}
    }
    % \vspace{-0.4cm}
    \caption{Data efficiency analysis of CAD. }
    \label{f6-2}
    % \vspace{-0.5cm}
\end{figure}

The augmentation operation would expand the size of the seed dataset, which could also contribute to OOD generalization. Therefore, to exclude the interference from data size and to further confirm the effectiveness of CAD as well as the ECF algorithm for improving OOD generalization capability, we collected the same size of original data to compare data efficiency, and the results are shown in Fig. \ref{f6-2}. Experimental results indicated that CAD generally outperformed the same size of original data (BiLSTM is the exception on the NLI task), while the ECF algorithm could steadily improve OOD generalization capability of CAD. % \uncertain{ok}

\begin{figure}[t]
    \centering
    \subfigure[original data]{
    \includegraphics[height=0.30\columnwidth]{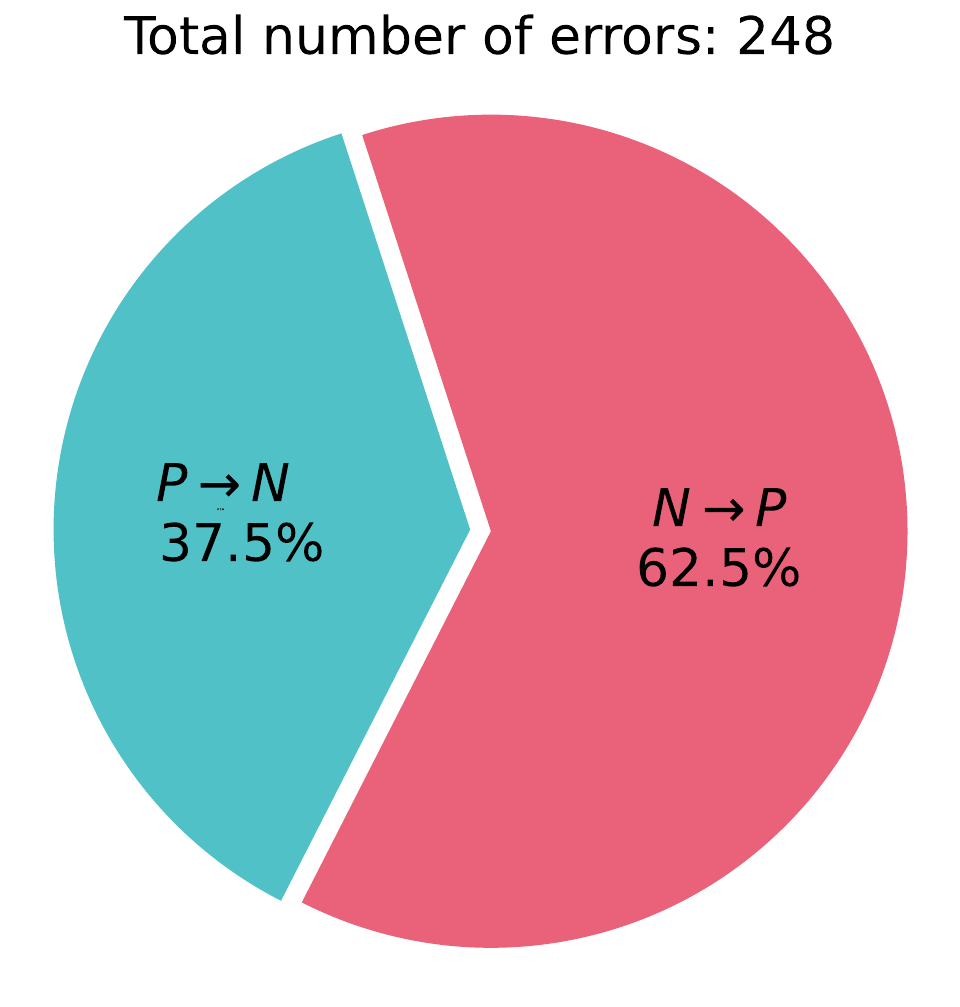} 
    \label{f6-3a}
    }
    \subfigure[CAD]{
    \includegraphics[height=0.30\columnwidth]{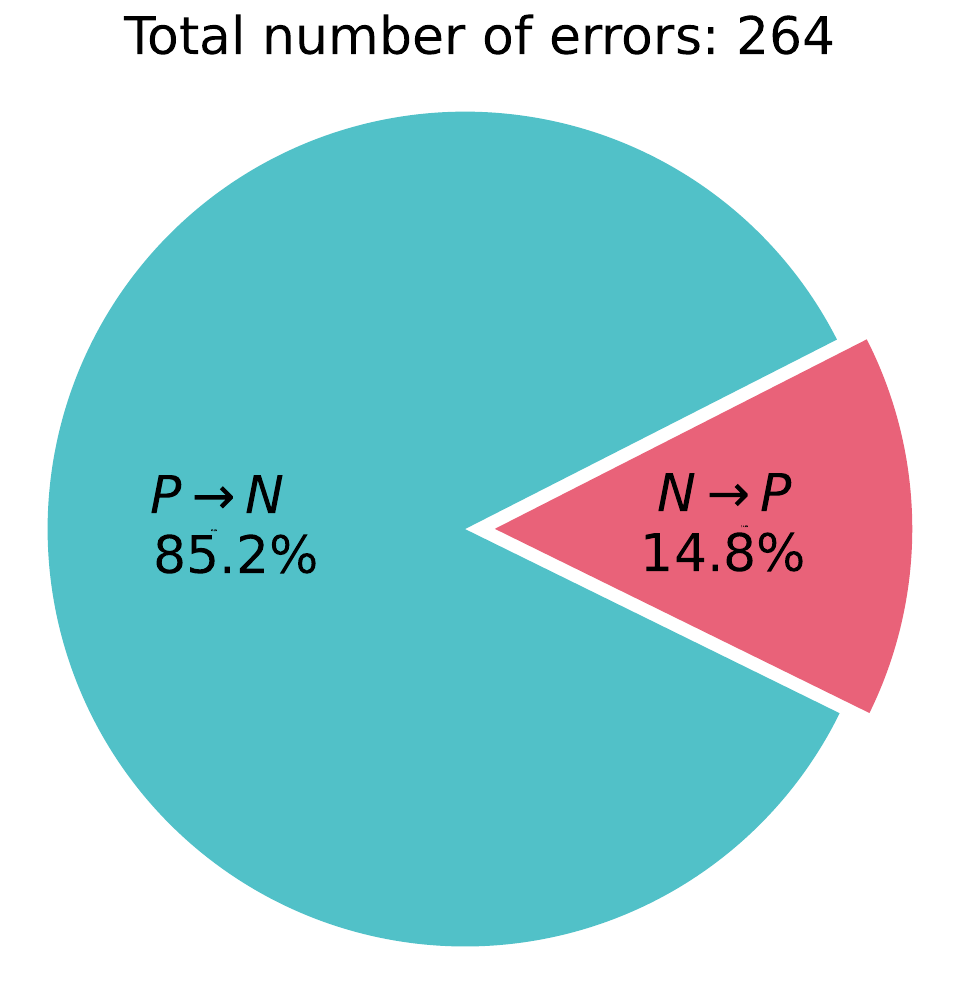} 
    \label{f6-3b}
    }
    \subfigure[CAD+ECF]{
    \includegraphics[height=0.30\columnwidth]{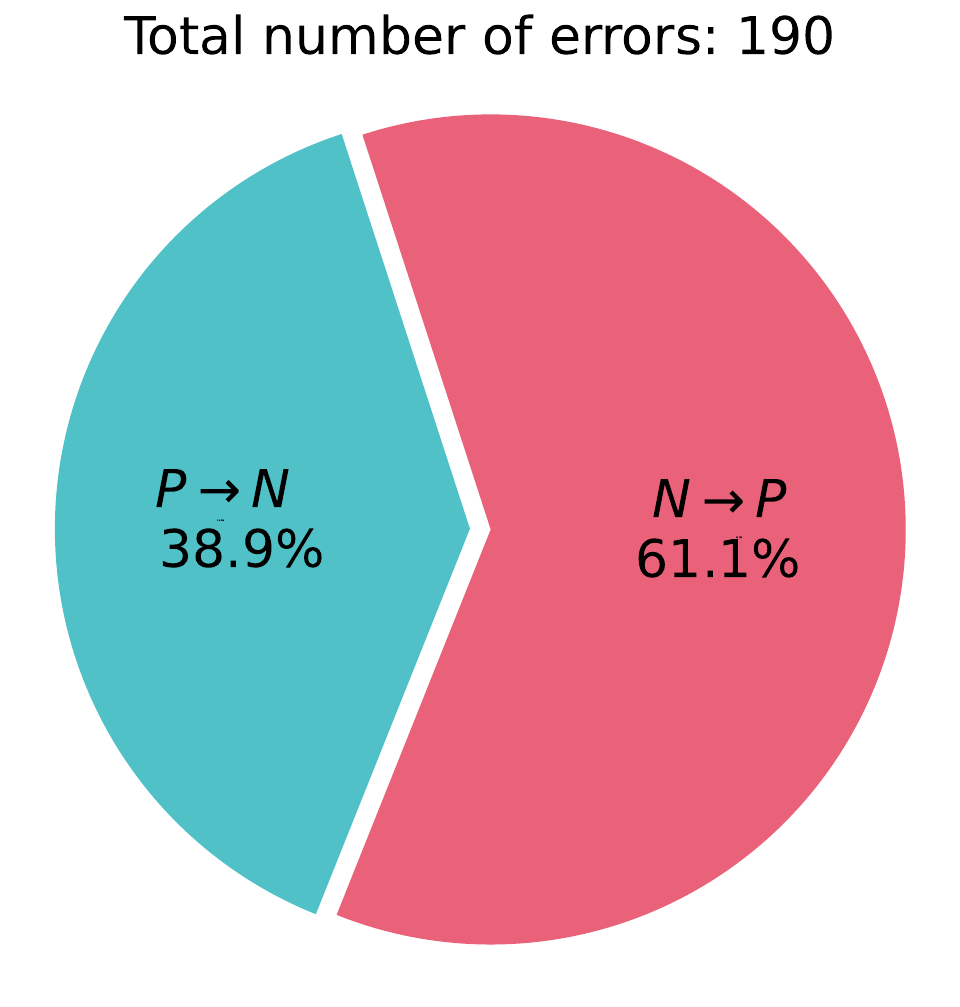} 
    \label{f6-3c}
    }
    \caption{Percentage of error case types in SA. The base model is Roberta and the OOD dataset is SST-2. $P \rightarrow N$ means positive cases that are predicted to be negative, and vice versa. }
    \label{f6-3}
\end{figure}

\begin{table}[t]
  \centering
%   \fontsize{5}{5}\selectfont
  \begin{threeparttable}
    % \resizebox{12.1cm}{!}{
    \begin{tabularx}{\textwidth}{cXcccc}
    \toprule
    \bf \# & \bf Sentence & \bf $Y$ & \bf $Y_{\text{ori}}$ & \bf $Y_{\text{CAD}}$ & \bf $Y_{\text{ECF}}$ \cr
    \midrule
    1 & \begin{scriptsize} it's \protect\sethlcolor{blue}\hl{predictable}, \protect\sethlcolor{pink}\hl{but} it \protect\sethlcolor{pink}\hl{jumps through} the \protect\sethlcolor{blue}\hl{expected hoops} with style and \protect\sethlcolor{pink}\hl{even} some \protect\sethlcolor{blue}\hl{depth}. \end{scriptsize} & P & P\begin{footnotesize}(0.99)\end{footnotesize} & N\begin{footnotesize}(0.52)\end{footnotesize} & P\begin{footnotesize}(0.78)\end{footnotesize} \cr
    \midrule
    2 & \begin{scriptsize} \protect\sethlcolor{pink}\hl{so} \protect\sethlcolor{blue}\hl{routine}, \protect\sethlcolor{blue}\hl{familiar} and \protect\sethlcolor{blue}\hl{predictable}, it \protect\sethlcolor{pink}\hl{raises the possibility} that it wrote \protect\sethlcolor{pink}\hl{itself as} a newly automated draft \protect\sethlcolor{blue}\hl{computer program}. \end{scriptsize} & N & N\begin{footnotesize}(0.87)\end{footnotesize} & P\begin{footnotesize}(0.51)\end{footnotesize} & N\begin{footnotesize}(0.52)\end{footnotesize} \cr
    \midrule
    3 & \begin{scriptsize} emerges \protect\sethlcolor{pink}\hl{as} something \protect\sethlcolor{blue}\hl{rare}, an \protect\sethlcolor{blue}\hl{issue movie} that's \protect\sethlcolor{pink}\hl{so} \protect\sethlcolor{blue}\hl{honest} and \protect\sethlcolor{blue}\hl{keenly observed} \protect\sethlcolor{pink}\hl{that} it \protect\sethlcolor{pink}\hl{doesn't} feel like one. \end{scriptsize} & P & P\begin{footnotesize}(0.93)\end{footnotesize} & N\begin{footnotesize}(0.79)\end{footnotesize} & P\begin{footnotesize}(0.81)\end{footnotesize} \cr
    \midrule
    4 & \begin{scriptsize} \protect\sethlcolor{pink}\hl{if} it tried to \protect\sethlcolor{blue}\hl{do anything more}, \protect\sethlcolor{pink}\hl{it would} \protect\sethlcolor{blue}\hl{fail} and perhaps \protect\sethlcolor{blue}\hl{explode}, \protect\sethlcolor{pink}\hl{but} at this level of \protect\sethlcolor{blue}\hl{manic whimsy}, it is \protect\sethlcolor{pink}\hl{just} about \protect\sethlcolor{blue}\hl{right}. \end{scriptsize} & P & P\begin{footnotesize}(0.99)\end{footnotesize} & N\begin{footnotesize}(0.63)\end{footnotesize} & P\begin{footnotesize}(0.85)\end{footnotesize} \cr
    \midrule
    5 & \begin{scriptsize} \protect\sethlcolor{pink}\hl{who knows} what \protect\sethlcolor{blue}\hl{exactly Godard is} on about in this film, \protect\sethlcolor{pink}\hl{but} his words and images \protect\sethlcolor{pink}\hl{don't have to} add up to \protect\sethlcolor{blue}\hl{mesmerize} you. \end{scriptsize} & P & P\begin{footnotesize}(0.94)\end{footnotesize} & N\begin{footnotesize}(0.80)\end{footnotesize} & P\begin{footnotesize}(0.62)\end{footnotesize} \cr
    \bottomrule
    \end{tabularx}
    % }
    \end{threeparttable}
    \caption{Five cases with complex sentence structure in SA. P/N denotes positive/negative respectively, and prediction probabilities are in parentheses. Emotional words are marked in \protect\sethlcolor{blue}\hl{blue} and structural words are marked in \protect\sethlcolor{pink}\hl{red}. }
    \label{tab6-1}
\end{table}

\subsection{Case Analysis}
\label{s7-4}

When counting the distribution of error cases in Fig. \ref{f6-3}, we observed an unexpected phenomenon: the language model trained on CAD was more likely to predict positive sentences as negative. Thus, we carefully compared the error cases of the language models trained on original data and CAD, and we found that the language model trained on CAD tended to make mistakes in sentences with complex sentence structures. The abnormal distribution in Fig. \ref{f6-3b} occurred because there were more positive examples with complex structures in the OOD dataset. We selected five typical cases with complex sentence structures in Table \ref{tab6-1} to illustrate this phenomenon concretely: Case 1 is a sentence with a transitive structure, linking two subparts with opposite meanings; Case 2 is a sentence with a progressive structure, where subparts have similar meanings; Cases 3, 4 and 5 have more complex sentence structures, including negative structure,  subjunctive mood, rhetorical structure, etc. 

Numerous cases illustrated that the language model trained on CAD failed to fully understand complex sentence structures, yet this situation did not occur with the language model trained on the original data. We consider this to be a concrete manifestation of the myopia phenomenon. As CAD is based on the principle of minimal editing \citep{Kaushik2020LearningTD}, the augmentation operation would focus more on the editing of words and ignore the changes to sentence structure \citep{Wang2021RobustnessTS,Yang2021ExploringTE}. According to our analysis in Section \ref{s4}, language models trained on CAD would therefore be more sensitive to words and less attentive to understanding sentence structure. However, sentence structure is not one of the correlation features, but could significantly influence the sentence meaning and is crucial for OOD generalization, so the neglect of sentence structure to some extent corresponds to the exclusion of the non-edited causal features in the myopia phenomenon. In addition, the ECF algorithm performed well in these cases, reflecting its capacity to alleviate the myopia phenomenon and unlock the potential of CAD. % \uncertain{ok}

\section{Limitations}

There were three main limitations in this study. Firstly, the assumptions made in Sections \ref{s3} and \ref{s4} about the distribution of sentence features and counterfactual augmentation operations were too idealistic. The feature extraction process we assumed for language models was likely to be oversimplified, and the effects of augmentation operations could be much more complex than we assumed. Secondly, we ignored the interaction between causal components, instead, we assumed that the feature components were independent of each other. We believed that this assumption required finer-grained adjustment. In addition, our study required high-quality CAD. However, there is currently a lack of CAD for complex tasks and reliable CAD construction techniques. % \uncertain{ok}

\section{Conclusion}

In this paper, we attributed the inefficiency of CAD in OOD generalization to the myopia phenomenon caused by counterfactual augmentation operations. We designed dataset-level and sentence-level constraints based on the structural properties of CAD to help language models to extract more complete causal features and then unlock the potential of CAD. 

Essentially, our study aims to design a more suitable algorithm (ECF algorithm) for specific dataset (CAD) to improve the OOD generalization capability of language models. Previous studies on OOD generalization often regarded data-based methods and algorithm-based methods as two separate technical paths. However, our proposed method seeks to fully combine the strengths of these two technical paths. With the advent of the era of large language models \citep{DBLP:conf/nips/BrownMRSKDNSSAA20}, we consider that potential research directions based on CAD include: automatically constructing CAD with large language models; CAD-guided demonstration selection in in-context learning; and applying CAD in the alignment process \citep{DBLP:journals/corr/abs-2203-02155} of large language models. 

\section*{Declaration of Competing Interest}

The authors declare that they have no known competing financial interests or personal relationships that could have appeared to influence the work reported in this paper. 

\section*{Acknowledgments}

This paper is a heavily extended version of our previously accepted conference paper from the 2023 IEEE International Conference on Acoustics, Speech, and Signal Processing (5 pages, including 1 page of references). Compared to the conference paper, this version analyzes the myopia phenomenon in feature space from the perspective of Fisher's Linear Discriminant and explains the theoretical basis for the additional constraints in the ECF algorithm. This version also adds approximately 50\% more experiments on an additional language model and provides a more detailed analysis of the experimental results, as well as a substantial extension of related work. 

This work was supported by the Shanghai Municipal Science and Technology Major Project (2021SHZDZX0102), and the Fundamental Research Funds for the Central Universities.

\clearpage

\bibliography{reference}

\end{document}